\documentclass{article}

\usepackage[preprint]{neurips_2019}

\newcommand\numberthis{\addtocounter{equation}{1}\tag{\theequation}}
\usepackage{neurips_2019}
\usepackage[utf8]{inputenc}
\usepackage[utf8]{inputenc} 
\usepackage[T1]{fontenc}    
\usepackage{hyperref}       
\usepackage{url}            
\usepackage{booktabs}       
\usepackage{amsmath, amsfonts,amsthm,amssymb}       
\usepackage{nicefrac}       
\usepackage{microtype}      
\usepackage{caption}
\usepackage{subcaption}
\usepackage{graphicx}
\newtheorem{proposition}{Proposition}
\usepackage[linesnumbered,ruled]{algorithm2e}
\usepackage{xcolor}
\usepackage{float}
\usepackage[skins]{tcolorbox}

\title{BreGMN: {scaled-Bregman Generative Modeling Networks}
} 

\author{%
  Akash Srivastava \\
  MIT-IBM Watson AI Lab\\
  IBM Research\\
  Cambridge, MA \\
  \texttt{akash.srivastava@ibm.com} \\
   \And
  Kristjan Greenewald \\
  MIT-IBM Watson AI Lab\\
  IBM Research\\
  Cambridge, MA \\
  \texttt{kristjan.h.greenewald@ibm.com} \\
   \AND
  Farzaneh Mirzazadeh \\
  MIT-IBM Watson AI Lab\\
  IBM Research\\
  Cambridge, MA \\
  \texttt{farzaneh@ibm.com} \\
}
\date{April 2019}

\begin{document}
\maketitle

\begin{abstract}
The family of $f$-divergences is ubiquitously applied to generative modeling in order to 
adapt
the distribution of the model to that of the data.  Well-definedness of $f$-divergences, however, requires the distributions of the data and model to overlap completely
in every time step of training. 
As a result, as soon as the support of distributions of data and model
contain non-overlapping portions, 
gradient-based training of the corresponding model becomes hopeless. 
Recent advances in generative modeling are full of remedies for handling this support mismatch problem: key ideas include either modifying the objective function to integral probability measures (IPMs) that are well-behaved even on disjoint probabilities, or optimizing a well-behaved variational lower bound instead of the true objective. We, on the other hand, establish that a complete change of the 
objective function is unnecessary, and instead an augmentation of the \emph{base measure} of the problematic divergence can resolve the issue. 
Based on this observation, we propose a generative model which leverages the class of \emph{Scaled Bregman Divergences} and 
generalizes both $f$-divergences and Bregman divergences. We analyze this class of divergences and show that with the appropriate choice of base measure it can resolve the support mismatch problem and incorporate geometric information.
Finally, we study the performance of the proposed method and 
demonstrate promising results on MNIST, CelebA and CIFAR-10 datasets. 
\end{abstract}

\maketitle
\section{Introduction}
\label{sec:intro}
Modern deep generative modeling paradigms offer a powerful approach for learning data distributions. 
Pioneering models in this family such as generative adversarial networks (GANs)  \citep{GoodfellowPMXWOCB14} and variational autoencoders (VAEs) \citep{KingmaW13} 
propose 
elegant solutions 
to generate high quality photo-realistic images, which were later evolved to generate other modalities of data. Much of the success of attaining photo-realism in generated images is attributed to the \textit{adversarial} nature of training 
in GANs. 
%
Essentially, GANs are  neural samplers 
in which a deep neural network $G_\phi$ is trained to generate high dimensional  samples  
from some low dimensional noise input. During the training, the generator is pitched against a classifier: the classifier is trained to distinguish the generated 
from the true data samples and the generator is simultaneously trained to generate samples that look like true data. Upon successful training,  the classifier fails to distinguish between the generated and actual samples. Unlike VAE, GAN is an implicit generative model since its likelihood function is implicitly defined and is in general intractable. Therefore training and inference are carried out using likelihood-free techniques such as the one described above.

In its original formulation, GANs can be shown to approximately minimize an $f$-divergence measure between the true data distribution $p_x$ and the distribution $q_\phi$ induced by its generator $G_\phi$. The difficulty in training the generator using the $f$-divergence criterion is that the supports of data and model distributions need to perfectly match. If at any time in the training phase, the supports have non-overlapping portions, the divergence either maxes out 
or becomes undefined. If the divergence or its gradient cannot be evaluated, it cannot, in turn, direct the weights of model towards matching distributions \citep{arjovsky2017wasserstein} and training fails. 

In this work, we present a novel method, 
BreGMN, for implicit adversarial and non-adversarial generative models that is based on \emph{scaled Bregman divergences} \citep{StummerV12} and does not suffer from the aforementioned problem of support mismatch. Unlike $f$-divergences, scaled Bregman divergences can be defined with respect to a base measure such that they stay well-defined even when the data and the model distributions do not have matching support.
Such an observation leads to a key contribution of our work, which is to identify base measures that can
play such a useful role. We find that measures whose support include the supports of data and model are the ones applicable.
In particular, we leverage Gaussian distributions to augment distributions of data and model into a base measure that guarantees the desired behavior. Finally we propose training algorithms for both adversarial and non-adversarial versions of the proposed model.

The proposed method facilitates a steady decrease of the objective function  and hence progress of training. We empirically evaluate the advantage of the proposed model for generation of synthetic and real image data. First,  we study simulated data in a simple 2D setting with mismatched supports and show the advantage of our method in terms of convergence.
Further, we evaluate BreGMN when used to train both adversarial and non-adversarial generative models.  For this purpose, we provide illustrative results on the MNIST, CIFAR10, and CelebA datasets, that show comparable performance to the sample quality of the state-of-art methods. In particular, our quantitative results on generative real datasets also demonstrate the effectiveness of the proposed method in terms of sample quality.

%
%
%
%
%
The remainder of this document is organized as follows. Section \ref{sec:related} outlines related work. We introduce the scaled Bregman divergence in Section \ref{sec:discrepancy}, demonstrate how it generalizes a wide variety of popular discrepancy measures, and show that with the right choice of base measure it can eliminate the support mismatch issue. Our application of the scaled Bregman divergence to generative modeling networks is described in Section~\ref{sec:method}, with empirical results 
presented in Section \ref{sec:results}. Section \ref{sec:conc} concludes the paper.
%
%
\section{Related work}\label{sec:related}
%
 Since the genesis of adversarial generative modeling,
there has been a flurry of work in this domain, e.g. \citep{NowozinCT16, SrivastavaVRGS17, mmdgan,arjovsky2017wasserstein} covering both practical and theoretical challenges in the field.  
Within this, a line of research 
addresses
the serious problem of \emph{support mismatch} 
that makes training hopeless if not remedied. One proposed way to alleviate this problem and stabilize training is to match the distributions of the data and the model based on a different, well-behaved discrepancy measure that can be evaluated even if the distributions are not equally supported. Examples of this approach include Wasserstein GANs \citep{arjovsky2017wasserstein} that replace the $f$-divergence with Wasserstein distance between distributions and other integral probability metric (IPM) based methods such as MMD GANs \citep{mmdgan}, Fisher GAN \citep{mroueh2017fisher}, etc. While IPM based methods are better behaved with respect to the non-overlapping support issue, they have their own issues. For example, MMD-GAN requires several additional penalties such as feasible set reduction in order to successfully train the generator. Similarly, WGAN requires some ad-hoc method for ensuring the Lipschitz constraint on the critic via gradient clipping, etc. 
Another approach to remedy the support mismatch issue comes from 
\cite{NowozinCT16}. 
They showed how GANs can be trained by optimizing a variational lowerbound to the actual $f$-divergence the original GAN formulation proposed. They also showed how the original GAN loss minimizes a Jenson-Shannon divergence and how it can be modified to train the generator using a $f$-divergence of choice. 

In parallel, works such as \citep{amari2010information} have studied the relation between the many different divergences available in the literature.
An important extension to \emph{Bregman} divergences, namely \emph{scaled Bregman divergences}, was proposed in the works of \cite{StummerV12,KisslingerS13} and generalizes both $f$-divergences and Bregman divergences.
The Bregman divergence in its various forms has long been used as the objective function for \emph{training} machine learning models. Supervised learning based on least squares (a Bregman divergence) is perhaps the earliest example. \cite{helmbold1995worst, auer1996exponentially, kivinen1998relative} study the use of Bregman divergences as the objective function for training single-layer neural networks for univariate and multivariate regression, along with elegant methods for matching the Bregman divergence with the network's nonlinear transfer function via the so-called \emph{matching loss} construct. 
In unsupervised learning, Bregman divergences are unified as the objective for  clustering in 
\cite{BanerjeeMDG05},
while convex relaxations of Bregman clustering models are proposed in \cite{ChengZS13}.
%
%
Generative modeling 
based on Bregman divergences is explored in  \cite{uehara2016b, 
uehara2016generative}, which relies on a duality relationship between Bregman and $f$ divergences. These works retain the $f$-divergence based $f$-GAN objective, but use a Bregman divergence as a distance measure for estimating the needed density ratios in the $f$-divergence estimator. This contrasts with our approach which uses the scaled Bregman divergence as the overall training objective itself.  



\section{Generative modeling via discrepancy measures}
\label{sec:discrepancy}
The choice of distance measure between the data and the model distribution is critical, as the success of the training procedure largely depends on the ability of these distance measures to provide meaningful gradients to the optimizer.
Common choices for distances include the Jensen-Shannon divergence (vanilla GAN) $f$-divergence ($f$-GAN) \citep{NowozinCT16} and various integral probability metrics (IPM, e.g. in Wasserstein-GAN, MMD-GAN) \citep{arjovsky2017wasserstein, mmdgan}. In this section, we consider 
a generalization of the Bregman divergence that also subsumes the Jensen-Shannon and $f$-divergences as special cases, and can be shown to incorporate some geometric information in a way analogous to IPMs.

\subsection{Scaled Bregman divergence}
The \textbf{Bregman divergence} \citep{bregman1967relaxation} forms a measure of distance between two vectors $p,q \in \mathbb{R}^d$ using a convex function $F: \mathbb{R}^d \rightarrow \mathbb{R}$ as
\begin{align*}
    B_F(p, q) = F(p) - F(q) - \nabla F(q) \cdot(p- q), 
\end{align*}
which includes a variety of distances, such as the squared Euclidean distance and the KL divergence between finite-cardinality probability mass functions, as special cases.

More useful in our setting is the class of \textbf{separable} Bregman divergences of the form
\begin{equation}
B_f(P,Q) = \int_{\mathcal{X}} f(p(x)) - f(q(x)) - f'(q(x)) (p(x) - q(x)) dx
\label{eq:sepBreg}
\end{equation}
where $f: \mathbb{R}^+\rightarrow \mathbb{R}$ is a convex function, $f'$ is its right derivative and $P$ and $Q$ are measures on $\cal X$ with densities $p$ and $q$ respectively.
In this form the divergence is a discrepancy measure for distributions as desired. 
In general, as the name divergence implies, the quantity is non-symmetric. It does not satisfy the triangle inequality either \citep{acharyya2013bregman}.

While this is a valid discrepancy measure, the Bregman divergence does not yield meaningful gradients for training when the two distributions in question have non-overlapping portions in their support, similar to the case of $f$-divergences \citep{ArjovskyB17}.
We thus propose to use the \emph{scaled} Bregman divergence, which introduces a third measure $M$ with density $m$ that can depend on $P$ and $Q$ and uses it as a base measure for the Bregman divergence. Specifically, the \textbf{scaled Bregman divergence} \citep{StummerV12} is given by
\begin{align}
B_f(P,Q |M) = \int_{\cal X} f\left(\frac{p(x)}{m(x)}\right)- f\left(\frac{q(x)}{m(x)}\right) -f'\left(\frac{q(x)}{m(x)}\right) \left(\frac{p(x)}{m(x)} - \frac{q(x)}{m(x)} \right) d M.
\label{eq:scaledBreg}
\end{align}
This expression is equal to the separable Bregman divergence \eqref{eq:sepBreg} when $M$ is equal to the Lebesgue measure. 

As shown in \citep{StummerV12}, the scaled Bregman divergence \eqref{eq:scaledBreg} contains many popular discrepancy measures as special cases. In particular, when $f(t) = t \log t$ it reduces to the \textbf{KL~divergence} for any choice of $M$  (as does the vanilla Bregman divergence).

Many classical criteria (including the KL and Jensen-Shannon divergences) belong to the family of \textbf{$f$-divergences}, defined as
\begin{align*}
    D_f(P, Q) = \int_{\cal{X}} q(x) f\left(\frac{p(x)}{q(x)}\right)dx.
\end{align*}
where the function $f : R_+ \to  R$ is a convex, lower-semi-continuous function satisfying $f(1) = 0$, where the densities $p$ and $q$ are absolutely continuous with respect to each other. The scaled Bregman divergence with choice of $M= Q$ reduces to the $f$ divergence family as:
\[
B_f(P,Q |Q) = \int_{\cal X} f\left(\frac{p(x)}{q(x)}\right) -f'\left(1\right) \left(\frac{p(x)}{q(x)} - 1 \right) d Q = \int_{\cal{X}} q(x) f\left(\frac{p(x)}{q(x)}\right)dx,
\]
which shows all $f$-divergences are special cases of the scaled Bregman divergence.
%
%
A more complete list of discrepancy measures included in the class of scaled Bregman divergences is found in \cite{StummerV12}.

\subsection{Noisy base measures and support mismatch}
A widely-known weakness of $f$-divergence measures is that when the supports of $p$ and $q$ are disjoint,
the value of the divergence is trivial or undefined. In the context of generative models, this issue is often tackled by adding noise to the model distribution which extends its support over the entire observed space such as in VAEs. However, adding noise to the observed space is not particularly well-suited for tasks such as image generation as it results in blurry images. In this work we propose choosing a base measure $M$ that in some sense incorporates geometric information in such a way that the gradients in the disjoint setting become informative without compromising the image quality. 

For the  scaled Bregman $B_f (P,Q |M)$, we propose choosing a ``noisy" base measure $M$, specifically one that is formed by convolving some other measure with the Gaussian measure $\mathcal{N}(0, \Sigma)$. Recall that convolution of two distributions corresponds to the addition of the associated random variables, hence in this case we are in affect adding Gaussian noise to the variable generated by $M$. In addition to adding noise, we require a base measure $\tilde{M}$ that depends on $P$ and $Q$ to avoid the vanilla Bregman divergence's lack of informative gradients (see Section \ref{sec:BregEst} below). By analogy to the Jensen-Shannon divergence, we choose
\begin{equation}\label{eq:baseMeasure}
\tilde M = \alpha (P \ast \mathcal{N}(0, \Sigma_1)) + (1-\alpha) (Q \ast {\cal N}(0,\Sigma_2  ))  
\end{equation}
for $0 \le \alpha\le 1$ and some covariances $\Sigma_1$ and $\Sigma_2$, where $\ast$ denotes the convolution of two distributions. Denote the density of $\tilde M$ as $\tilde m$.

Importantly, observe that each term of the corresponding scaled Bregman $B_f (P,Q |\tilde M)$ is always well defined and finite (with the exception of certain choices of $f$ such as $-\log$ that require numerical stabilization similar to the case of $f$-divergence) since $\tilde M$ has full support. 
Furthermore, since $\tilde M$ is a noisy copy of $\alpha P + (1-\alpha) Q$, the ratio $\frac{p}{\tilde m}$ will be affected by $q$ even outside the support of $q$, and vice versa. This ensures that a training signal remains in the support mismatch case.

The presence of this training signal seems to indicate that geometric information is being used, since it varies with the distance between the supports.
To further explore this intuitive connection between noisy base measures and geometric information, we attempt to relate $B_f(P,Q|\tilde{M})$ to the $W_p$ distance. In what follows, for simplicity we focus on the case of $f(t) = t \log t$; analysis for more general choices of $f$ is left for future work. For the KL divergence for example, Pinsker's inequality states that 
\[
D_{KL}(p||q) \geq 2(W_1(p,q))^2 .
\]
A similar lower bound for the $W_2$ distance and certain log-concave $q$ follows from Talagrand's inequality \citep{bobkov2000brunn}. These lower bounds are not surprising, since the KL divergence can go to infinity when Wasserstein-p is finite. 
However, lower bounds of this type are not sufficient to imply that a divergence is using geometric information, since it can increase very quickly while $W_p$ increases only slightly.

Our use of a noisy $M_0$, however, allows us to obtain an upper bound for a symmetrized version of $B_f(P,Q|\tilde{M})$, which implies a continuity with respect to geometric information. While we found in our generative modeling experiments that a symmetrized version is unnecessary to use in practice, it is useful for comparison to IPMs. 
Recall that the Jensen-Shannon divergence 
constructs a symmetric measure by symmetrizing the KL divergence around $(P+Q)/2$. Any Bregman divergence can be similarly symmetrized (Eq. 16 in \cite{nielsen2011skew}). 
For simplicity, we consider the special case of $\tilde{M}$, namely $M_0 = \frac{{P} + {Q}}{2}\ast \mathcal{N}_\sigma$ with density $m_0$, and use it to both scale and symmetrize the scaled Bregman divergence, obtaining the measure 
$
{B_f(P,M_0 |M_0) + B_f(Q,M_0 |M_0)} 
= {D_f(P || M_0) + D_f(Q||M_0)}$. 
In Section \ref{supp:wass} of the Supplement we prove:
\begin{proposition}\label{prop:wass}
Assume that $\mathbb{E}_{U\sim P} \|U\|$ and $\mathbb{E}_{V\sim P} \|V\|$ are bounded. Then
\[
\left|B_{t \log t}(P,M_0 |M_0) - B_{t\log t}(Q,M_0 |M_0)\right| \leq c W_2(P,Q) + |h(Q) - h(P)|,
\]
where $c$ is a constant given in the proof and $h(P)$ is the Shannon entropy of $P$. 
\end{proposition} 

While an $h(P) - h(Q)$ term remains, it is simple to rescale $Q$ to match the entropy of $P$, eliminating that term and leaving the Wasserstein distance.\footnote{Under certain smoothness conditions on $P$ and $Q$ $|h(P) - h(Q)|$ can itself be upper bounded by the Wasserstein distance (see \cite{Polyanskiy016a} for details).} 

While not fully characterizing the geometric information in $B_f(P,Q|M_0)$, these observations seem to imply that the use of the noisy $\tilde{M}$ is capable of incorporating some geometric information without having to resort to IPMs with their associated training difficulties in the GAN context such as gradient clipping and feasible set reduction \citep{ArjovskyB17,mmdgan}.

\section{Model}
\label{sec:method}

Let $\{x_i \vert x_i \in \mathbb{R}^d\}_{i=1}^N$ be a set of $N$ samples drawn from the data generating distribution $p_x$ that we are interested in learning through a parametric model $G_\phi$. 
The goal of generative modeling is to train $G_\phi$, 
generally implemented as a deep neural network, to map samples from a $k$-dimensional easy-to-sample distribution 
to the ambient $d$ dimensional data space, i.e. $G_\phi:\mathbb{R}^k \mapsto \mathbb{R}^d$. Letting $q_\phi$ be the distribution induced by the generator function $G_\phi$, almost all training criteria are of the form
\begin{align}
    \label{eq:gen_loss}
    \min_\phi 
    {D}(p_x \Vert q_\phi)
\end{align}
where $\mathbb 
{D(.\Vert.)}$ is 
a measure of discrepancy between the data and the model distributions. 
We propose to use the 
{scaled-Bregman divergence} as 
${D}$ in Equation \eqref{eq:gen_loss}. 
We will show 
that unlike $f$-divergences, scaled-Bregman divergences can be easily estimated
with respect to a base measure using only samples from the distributions. 
This is important when we aim to match distributions in very high dimensional spaces where they may not have any overlapping support \citep{arjovsky2017wasserstein}.  

In order to compute the divergence between data and model distributions, it is not required that both densities are known or can be evaluated on realizations from distributions. Instead, being able to evaluate the ratio between them, i.e. \emph{density ratio estimation}, is typically all that is needed. For example, generative models based on $f$-divergences only need density ratio estimation. Importantly, similar to the case of $f$-divergences, scaled-Bregman divergence estimation requires estimates of the density ratios only.  

Below, we describe two methods of 
density ratio estimation (DRE) between two distributions. 
In what follows, suppose $r=\frac{p_x}{q_\phi}$ is the density ratio.

\paragraph{Discriminator-based DRE:}
This family of models uses a discriminator to estimate the density ratio.
Let $y=1$ if $x\sim p_x$ and $y=0$ if $x\sim q_\phi$. Further, let $\sigma(C(x)) = p(y=1\vert x)$, namely the discriminator, be a trained binary classifier on samples from $p_x$ and $q_\phi$ where $\sigma$ is the Sigmoid function. 
It is then easy to show that $C(x) = -\log \frac{p_x(x)}{q_\phi(x)}= -\log r(x)$ \citep{dre}, so $C$ is a function of density ratio $r(x)$. 
In fact, this is the underlying principle in adversarial generative models \citep{GoodfellowPMXWOCB14}. As such, most discriminator-based DREs result in adversarial training procedures when used in generative models. 

\paragraph{MMD-based DRE:} 
This family of models estimate the density ratio without the use of a discriminator.
In order to estimate the density ratio $r$ without training a classifier, thereby avoiding adversarial training of the generator later, we can employ the maximum mean discrepancy (MMD) \citep{mmd} criterion as in \cite{dre}. 
By solving for $r$ in the RKHS in
\begin{align}
\label{eq:mmd-ratio}
\min_{r\in\mathcal{H}} \bigg \Vert \int k(x; .)p_x(x) dx - \int k(x; .)r(x)q_\phi(x) dx \bigg \Vert_{\mathcal{H}}^2,
\end{align}
where $k$ is a kernel function, we obtain a closed form estimator of the density ratio as
\begin{align}
\label{eq:ratio}
\hat{\mathbf{r}}_{p/q} = K_{q,q}^{-1} K_{q,p} \pmb 1. 
\end{align}
Here $K_{q,q}$ and $K_{q,p}$ denote the 
Gram matrices 
corresponding to kernel $k$.

\subsection{Empirical estimation}\label{sec:BregEst}
Using the DRE estimators introduced above we create empirical estimators of the scaled-Bregman divergence \eqref{eq:scaledBreg} as
\begin{align}
    \label{eq:bre_e}
    \hat{B}_f(p_x,q_\phi|M) = \frac{1}{N}\sum_{i=1}^N f\left(r_{p/m}(x_i)\right)- f\left(r_{q_\phi/m}(x_i)\right) \\ \nonumber 
    -f'\left(r_{q_\phi/m}(x_i)\right) \left(r_{p/m}(x_i) - r_{q_\phi/m}(x_i) \right)
\end{align}
where $r_{p/m}$ denotes a DRE of $p/m$ and the $x_i$ are i.i.d. samples from the base distribution $m$ with measure $M$.
Note that this empirical estimator $\hat{B}_f$ does not have gradients with respect to $\phi$ if we only evaluate the DRE estimators on samples from the base measure $m$. 
Choices of $M$ that depend on $p$ and $q$, however, including our choice of $\tilde M$ \eqref{eq:baseMeasure} as well as the choice $M = Q$ ($f$-divergences), have informative gradients, allowing us to train the generator. 

\subsection{Training}

Training  the generator function  $G_\phi$ using scaled-Bregman divergence (shown in Algorithm \ref{alg:training}) alternates the following two steps until convergence. 

\paragraph{Step 1:} Estimate the density ratios $r_{p/m}$ and $r_{q_\phi/m}$ using either the adversarial discriminator-based method (as in a GAN) or the non-adversarial MMD-based method.

\paragraph{Step 2:} Train the generator by optimizing
\begin{align}
\label{eq:train}
\min_\phi \hat{B}_f(p_x,q_\phi|\tilde M).
\end{align}




\begin{algorithm}[!t]
\SetAlgoNoLine
\While{ not converged }{ 
     \textbf{Step 1 } Estimate the density ratios $r_{p/m}$ and $r_{q_\phi/m}$,
     using either the adversarial discriminator-based (GAN-like) or\\ non-adversarial two-sample-test-based (MMD-like) method. \\
     \textbf{Step 2 } Train the generator by optimizing 
\\
     ${\hspace{6.3cm}\min_\phi \hat{B}_f(p_x,q_\phi|\tilde M)}$.
  }
\caption{Training Algorithm of BreGMN}
\label{alg:training}
\end{algorithm}

\section{Experiments}
\label{sec:results}

In this section we present a detailed evaluation of our proposed scaled-Bregman divergence based method for training generative models. Since most generative models aim to learn the data generating distribution, our method can be generically applied to a large number of simple or complex and deep generative models for training. We demonstrate this by training a range of simple to complex models with our method.

\subsection{Synthetic data: support mismatch}

\begin{figure}[!tb]
\centering
 \caption{$f$-divergence and scaled-Bregman divergence based training on synthetic dataset of two disjoint, non-overlapping 2D distributions.
 \label{fig:syn}}
\begin{subfigure}{0.5\textwidth}
  \includegraphics[width=\linewidth]{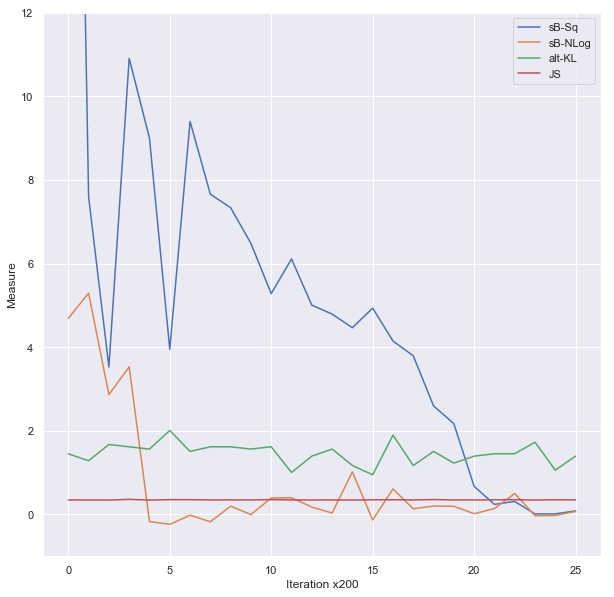}
  \caption{Measure vs Training}
\end{subfigure} 
\begin{subfigure}{0.48\textwidth}
  \includegraphics[width=\linewidth]{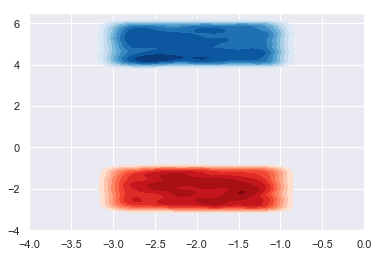}
  \caption{2D distributions.}
\end{subfigure}
\end{figure}

In this experiment, we evaluate our method in the regime where $p$ and $q_\phi$ have mismatched support, in order to validate the intuition that the noisy base measure $\tilde M$ aids learning in this setting. As shown in Figure \ref{fig:syn}(b), we start by training a simple probabilistic model (blue) to match the data distribution (red). The data distribution is a simple uniform distribution with finite support. Our model is a therefore parameterized as a uniform distribution with one trainable parameter. 

Figure \ref{fig:syn}(a) shows the effect of training this model with $f$-divergence and with our method. Clearly, neither the KL nor JS divergences are able to provide any meaningful gradients for the training of this simple model. Our scaled-Bregman based training method, however, is indeed able to learn the model. Interestingly, as Figure \ref{fig:syn} shows, the choice of the function $f$ matters in the empirical  convergence rate of our method, with the convergence of $f(t) = -\log t$ much faster than that of $f(t) = t^2$. 

\subsection{Non-adversarial generative model}
Our training procedure is not intrinsically adversarial, i.e. it is not a saddle-point problem when the MMD-based DRE is used. To demonstrate the capability of the proposed model in training non-adversarial models, in this section, we 
apply 
the MMD-based DRE to train a generative model on the MNIST dataset in a non-adversarial fashion. As shown in Figure \ref{fig:mnist}(a), our method can be used to successfully train generative models of a simple dataset without using adversarial techniques. While the sample quality is not optimal (better sample quality may be achievable by carefully tuning the kernel in the MMD criterion), the training procedure is remarkably stable as shown in Figure \ref{fig:mnist}(b). 

\begin{figure}[!tb]
\centering
 \caption{Non-adversarial Training using scaled-Bregman Divergence and MMD based DRE.
 \label{fig:mnist}}
\begin{subfigure}{0.48\textwidth}
  \includegraphics[width=\linewidth]{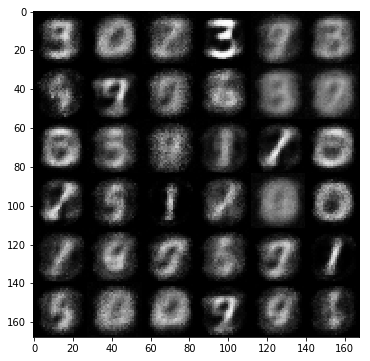}
  \caption{Samples from the generator.}
\end{subfigure} 
\begin{subfigure}{0.5\textwidth}
  \includegraphics[width=\linewidth]{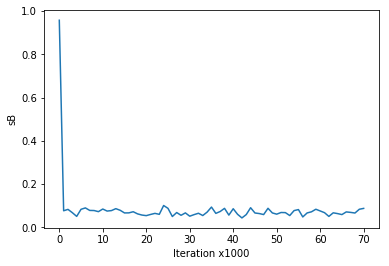}
  \caption{Generator loss steadily decreases.}
\end{subfigure}
\end{figure}

\subsection{Adversarial generative model}

\begin{figure}[!tb]
\centering
 \caption{Random samples from Adversarial BreGMN models (after 5 Epochs)
 \label{fig:mnist}}
\begin{subfigure}{0.48\textwidth}
  \includegraphics[width=\linewidth]{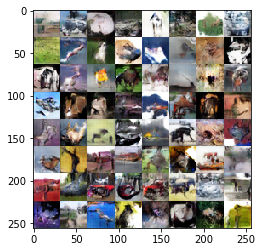}
  \caption{CIFAR10}
\end{subfigure} 
\begin{subfigure}{0.5\textwidth}
  \includegraphics[width=\linewidth]{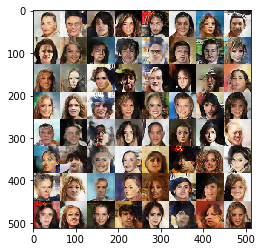}
  \caption{CELEB A}
\end{subfigure}
\end{figure}

Training generative models on complicated high-dimensional datasets such as those of natural images is done preferably with adversarial techniques since they tend to lead to better sample quality. One straightforward way to assign adversarial advantage to our method is to use a discriminator based DRE. To evaluate our training method on adversarial generation, in this section, we compare the Frechet Inception Distance (FID) \citep{fid} of MMD-GAN \citep{mmdgan}, GAN \citep{GoodfellowPMXWOCB14} against BreGMN on CIFAR10 and CelebA dataset. FID measures the distance between the data and the model distributions by embedding their samples into a certain higher layer of a pre-trained Inception Net. We used a 4-layer DCGAN \citep{dcgan} architecture for all the experiments and averaged the FID over multiple runs. $\mathcal{N}(0,0.001)$ is used as the noise level across all the experiments. MMD-GAN trains a generator network using the maximum mean discrepency \citep{mmd} where the kernel is trained in an adversarial fashion. As shown in Table \ref{tab:celeb}, both BreGMN and GANs performs better than MMD-GAN in terms of sample quality. While BreGMN performs slightly better than GAN on average, their sample qualities are comparable.

\begin{table}[!tb]
\centering
\caption{Sample quality (measured by FID; lower is better) of BreGMN compared to GANs. }
\label{tab:celeb}
\begin{tabular}{@{}lcccc@{}}
\toprule
\multicolumn{1}{c}{\textbf{Archtitecture}}  & \textbf{Dataset}                      & \textbf{MMD-GAN}                   & \textbf{GAN}                      & \textbf{BreGMN}                               \\ \midrule
\multicolumn{1}{l}{\textbf{DCGAN}}        & \multicolumn{1}{c}{\textbf{Cifar10}} & \multicolumn{1}{c}{40 }     & \multicolumn{1}{c}{26.82 } & \multicolumn{1}{c}{\textbf{26.62}} \\ 

\multicolumn{1}{l}{\textbf{DCGAN}}        & \multicolumn{1}{l}{\textbf{CelebA}}  & \multicolumn{1}{c}{41.10} & \multicolumn{1}{c}{30.97 } & \multicolumn{1}{c}{\textbf{30.84}} \\ \bottomrule
\end{tabular}
\end{table}

\section{Conclusions}
\label{sec:conc}
In this work, we proposed scaled-Bregman divergence based 
generative models and identified base measures for them to facilitate effective training.
%
We showed that the proposed approach provides a  certifiably advantageous
criterion 
to model the data distribution using deep generative networks in comparison to the $f$-divergence based training methods. We clearly established that unlike $f$-divergence based training our method does not fail to train even when the model and the data distributions do not have any overlapping support to start with. 
%
A future direction of research addresses the choice of the base measure  and the effect of noise level on the optimization. Another, more theoretical direction is to study and establish the relationship between scaled-Bregman divergence and other IPMs.




\bibliography{Breg_GAN_NeuRIPS2019.bib}
\bibliographystyle{icml2019}

\clearpage

\section*{\LARGE{Supplementary material for: BreGMN: scaled-Bregman Generative Modeling Networks}}
\appendix

\section{Proof of Proposition \ref{prop:wass}}\label{supp:wass}
Observe that
\begin{align*}
    &B_{t \log t}(P,M_0 |M_0) - B_{t\log t}(Q,M_0 |M_0)= D_{KL}(P || M_0) - D_{KL}(Q||M_0)\\
    &= \int_{\cal X} \log\left(\frac{p(x)}{m_0(x)}\right) dP - \int_{\cal X} \log\left(\frac{q(x)}{m_0(x)}\right) dQ\\
    &= \int_{\cal X} \log\left(m_0(x)\right) dQ - \int_{\cal X} \log\left(m_0(x)\right) dP + h(Q) - h(P) \\
    &= \mathbb{E}_{V \sim Q} \log\left(m_0(V)\right) - \mathbb{E}_{U \sim P} \log\left(m_0(U)\right)+ h(Q) - h(P) 
\end{align*}
where we denote the Shannon entropy as $h(P) = -\int_{\cal X} \log(p(x)) dP$. Note that
\begin{align*}
&|\log\left(m_0(V)\right) - \log\left(m_0(V)\right)| = \left|\int_0^1 \langle \nabla \log m_0(tv + (1-t)u), u-v \rangle dt\right|\\
&\leq \int_0^1\left(\frac{3}{\sigma^2} (t\|v\| + (1-t)\|u\|) + \frac{4}{\sigma^2}\left(\mathbb{E}_{U\sim P} \|U\| + \mathbb{E}_{V\sim Q} \|V\|\right)\right) \|u - v\|dt\\
&=\left(\frac{3}{2\sigma^2} (\|v\| + \|u\|) + \frac{4}{\sigma^2}\left(\mathbb{E}_{U\sim P} \|U\| + \mathbb{E}_{V\sim Q} \|V\|\right)\right) \|u - v\|\numberthis \label{eq:wassDecomp}
\end{align*}
where we have used Cauchy-Schwartz inequality and have noted that
\[
\|\nabla \log m_0(x)\| \leq \frac{3}{\sigma^2} \|x\| + \frac{4}{\sigma^2}\left(\mathbb{E}_{U\sim P} \|U\| + \mathbb{E}_{V\sim Q} \|V\|\right), \qquad \forall x \in \mathbb{R}^d,
\]
by Proposition 2 of \cite{Polyanskiy016a}. 

Let $W_p(\cdot,\cdot)$ denote the Wasserstein-$p$ distance
\[
W_p(\mu,\nu) := \left(\inf_{\pi \in \Pi(\mu,\nu)} \int_{\mathcal{X}\times  \mathcal{X}} \|x - y\|^p d\pi(x,y)\right)^{\frac{1}{p}},
\]
where $\Pi(\mu,\nu)$ denotes the set of \emph{couplings} of $\mu$ and $\nu$, i.e. the set of measures on $\mathcal X \times \mathcal X$ with marginals $\mu$ and $\nu$.

Now, taking the expectation of \eqref{eq:wassDecomp} with respect to the $W_2$-optimal coupling $\pi$ between $P$ and $Q$, we have
\begin{align*}
    &|B_{t \log t}(P,M_0 |M_0) - B_{t\log t}(Q,M_0 |M_0)|\\
    &\leq \mathbb{E}_{(u,v)\sim \pi} \left[\left(\frac{3}{2\sigma^2} (\|v\| + \|u\|) + \frac{4}{\sigma^2}\left(\mathbb{E}_{U\sim P} \|U\| + \mathbb{E}_{V\sim Q} \|V\|\right)\right) \|u - v\|\right] + |h(Q) - h(P)|\\
    &\leq \sqrt{\left(\mathbb{E}_{\pi} \left(\frac{3}{2\sigma^2} (\|v\| + \|u\|) + \frac{4}{\sigma^2}\left(\mathbb{E}_{U\sim P} \|U\| + \mathbb{E}_{V\sim Q} \|V\|\right)\right) \right) \left(\mathbb{E}_{\pi} \|u - v\|^2\right) }+ |h(Q) - h(P)|\\
    &= c W_2(P,Q) + |h(Q) - h(P)|,
\end{align*}
where we have again used the Cauchy-Schwarz inequality and have set the constant $c=  \frac{11}{2\sigma^2}\left(\mathbb{E}_{U\sim P} \|U\| + \mathbb{E}_{V\sim Q} \|V\|\right)$. \qed 

\end{document}